\documentclass[twoside,leqno,twocolumn]{article}

\usepackage[letterpaper]{geometry}

\usepackage{ltexpprt}

\usepackage[utf8]{inputenc} 
\usepackage[T1]{fontenc}    
\usepackage{hyperref}       
\usepackage{url}            
\usepackage{booktabs}       
\usepackage{amsfonts}       
\usepackage{nicefrac}       
\usepackage{microtype}      

\usepackage{graphicx}

\usepackage{algorithm}
\usepackage{algorithmic}
\usepackage{multirow}
\usepackage{float}
\usepackage{subfig}
\usepackage{amsmath} 
\usepackage{amssymb}
\usepackage{pifont}
\newcommand{\xmark}{\ding{55}}%
\usepackage{color}
\usepackage{wrapfig}
\usepackage{tikz}
\usetikzlibrary{shapes,arrows}
\usetikzlibrary{external}
\usetikzlibrary{positioning,calc}
\tikzset{%
  block/.style    = {draw, thick, rectangle, minimum height = 3em,
    minimum width = 3em},
  sum/.style      = {draw, circle, node distance = 2cm}, 
  input/.style    = {coordinate}, 
  output/.style   = {coordinate} 
}

\begin{document}

\title{Fisher Deep Domain Adaptation}
\author{Yinghua Zhang \thanks{Department of Computer Science and Engineering, Hong Kong University of Science and Technology (\{yzhangdx, yqsong, qyang\}@cse.ust.hk)}
, Yu Zhang \thanks{Department of Computer Science and Engineering, Southern University of Science and Technology (yu.zhang.ust@gmail.com)}
, Ying Wei \thanks{Tencent (judyweiying@gmail.com, kunbai@tencent.com)}
, Kun Bai \footnotemark[3] 
, Yangqiu Song \footnotemark[1]
, Qiang Yang \footnotemark[1] \thanks{WeBank}
}
\date{}


\maketitle


\fancyfoot[R]{\scriptsize{Copyright \textcopyright\ 2020 by SIAM\\
Unauthorized reproduction of this article is prohibited}}





\begin{abstract} \small\baselineskip=9pt 
Deep domain adaptation models learn a neural network in an unlabeled target domain by leveraging the knowledge from a labeled source domain. This can be achieved by learning a domain-invariant feature space. Though the learned representations are separable in the source domain, they usually have a large variance and samples with different class labels tend to overlap in the target domain, which yields suboptimal adaptation performance. To fill the gap, a Fisher loss is proposed to learn \emph{discriminative} representations which are within-class compact and between-class separable. Experimental results on two benchmark datasets show that the Fisher loss is a general and effective loss for deep domain adaptation. Noticeable improvements are brought when it is used together with widely adopted transfer criteria, including MMD, CORAL and domain adversarial loss. For example, an absolute improvement of $6.67\%$ in terms of the mean accuracy is attained when the Fisher loss is used together with the domain adversarial loss on the \textit{Office-Home} dataset. 
\end{abstract}

\section{Introduction}

The success of deep neural networks usually relies on massively labeled data. Unfortunately, labeling involves tremendous human efforts and costs. To reduce the dependency on labeled data, domain adaptation models are developed for learning in an unlabeled \emph{target} domain with the help of a \emph{source} domain where abundant labeled data are available \cite{pan2010survey,weiss2016survey}. 
One widely adopted approach is to find a domain-invariant latent feature space. Such a space can be found by minimizing the discrepancy between the two domains which can be estimated by various measures \cite{gretton_mmd_nips_2012,sun_coral_eccv_2016,ganin_dann_jmlr_2016}. After the latent feature space is found, samples from the two domains are projected into that space and a task network is learned with the labeled source domain data. Since the distributions of the two domains are similar in the resultant feature space, it is expected that the task network built in the source domain can generalize to the target domain. 

\begin{figure*}[!htbp]
    \centering
    \subfloat[w/o the Fisher loss] {
    \begin{minipage}[c]{0.225\linewidth}
    \centering
    \includegraphics[scale=0.4]{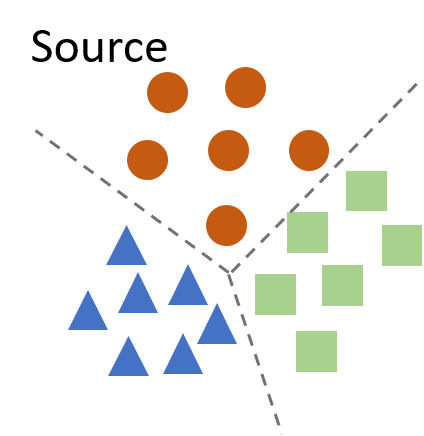}
    \label{fig:naive_source_domain}
    \end{minipage}
    \hfill
    \begin{minipage}[c]{0.225\linewidth}
    \centering
    \includegraphics[scale=0.4]{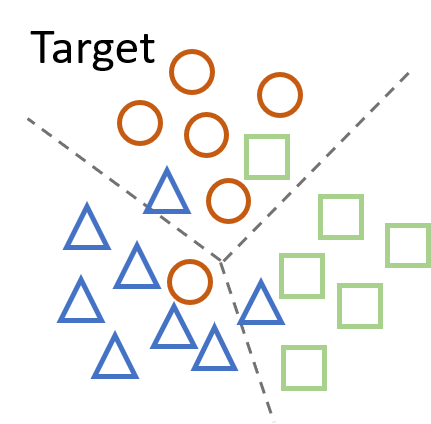}
    \label{fig:naive_target_domain}
    \end{minipage}
    }
    \hfill
    \subfloat[w/ the Fisher loss] {
    \begin{minipage}[c]{0.225\linewidth}
    \centering
    \includegraphics[scale=0.4]{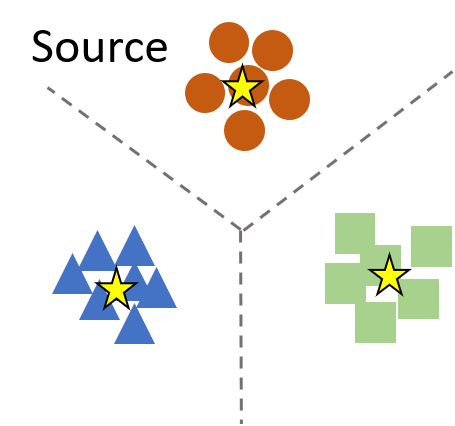}
    \label{fig:fisher_source_domain}
    \end{minipage}
    \hfill
    \begin{minipage}[c]{0.225\linewidth}
    \centering
    \includegraphics[scale=0.4]{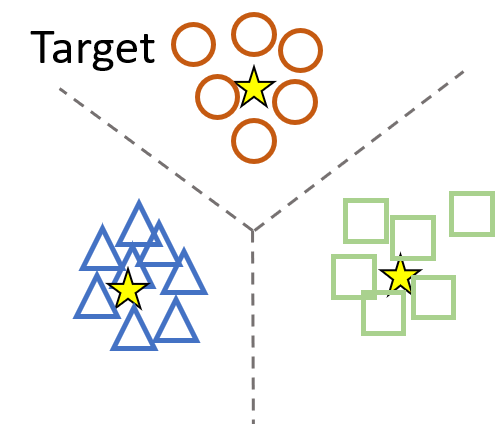}
    \label{fig:fisher_target_domain}
    \end{minipage}
    }
    \caption{A toy example. With the Fisher loss, the features in the two domains are within-class compact and between-class separable, making the hypothesis generalize to the target domain. (Samples with the same class label are plotted with the same color. Class centers are denoted by yellow stars. The decision boundary is plotted as gray dashed lines. )}
    \label{fig:toy_example}
\end{figure*}

Though source domain samples are separable and distributions of the two domains are similar, misclassifications are observed when an ideal classifier in the source domain is directly applied in the target domain. This is because there is a large variance within the source domain features and samples from different classes overlap with each other in the target domain. A toy example is shown in Fig. \ref{fig:toy_example}. 

This motivates an additional Fisher loss, which mitigates the aforementioned phenomenon by enforcing the feature discriminability. The discriminability requires two properties: within-class compactness and between-class separability. The Fisher loss introduces a center for each class and estimates the centers with the labeled source domain data. In the source domain where the class labels are known, the samples with the same class label are pushed close to the corresponding class center while the centers of different classes are pulled away. Consequently, the features in the source domain become discriminative. Meanwhile, as the domain discrepancy between the two domains is minimized, the features of the target domain are optimized to follow the distribution of the source domain features and they become discriminative as well. Though the label/center assignment is unknown in the target domain, the discriminability can be further enhanced under the guidance of the entropy minimization regularization \cite{grandvalet_entropy_minimization_nips_2005} which enforces a target domain sample to move towards one of the centers. When the adaptation is completed, as shown in Fig. \ref{fig:fisher_target_domain}, the features of both domains are discriminative and the decision boundary generalizes to the target domain well. 
Two instantiations of the Fisher loss, which are motivated by Fisher's discriminant analysis and Maximum Margin Criterion (MMC) \cite{li_mmc_nips_2004}, are developed. Experimental results demonstrate that the Fisher loss is a general and effective loss for deep domain adaptation, as it boosts the adaptation performance of three popular transfer criteria, including MMD, deep CORrelation ALignment (CORAL), and domain adversarial loss. Moreover, with this simple loss, basic transfer methods can perform comparably or even better than state-of-the-art models. 

\section{Related Works}


To address the dilemma between the limited training data and the data-hungry nature of machine learning models, transfer learning has been extensively studied \cite{pan2010survey,weiss2016survey}. One popular transfer learning approach is feature-based transfer learning which learns a latent feature space shared by the two domains. Such shared feature space can be found by minimizing a distance measure of two distributions, for example, the MMD \cite{pan_tca_2011,tzeng_ddc_arxiv_2014,wei_l2t_icml_18}. In addition to such statistical distances, the domain discrepancy can also be parameterized as a neural network and the common feature space can be found by optimizing an adversarial objective \cite{ganin_dann_jmlr_2016,tzeng_adda_cvpr_2017,li_ijcai_2017}. In this approach, domain-invariant features are learned by minimizing the training loss in the source domain while maximizing the loss of distinguishing the two domains. 


Classical dimension reduction techniques include principal component analysis, Linear Discriminant Analysis (LDA), non-negative matrix factorization, etc. These techniques embed high-dimensional data into a low-dimensional space via linear transformations or kernel tricks. As deep learning has risen as a new paradigm for representation learning, learning discriminative features with deep neural networks has drawn attention in recent years. A contrastive loss is developed in \cite{sun_deepid2_nips_2014}. Given a pair of faces, the representations of faces with the same identity should be close to each other. Otherwise, their distances should exceed a large threshold. A triplet loss is used in \cite{schroff_facenet_cvpr_2015} which trains on triplets of matching/non-matching face patches. The distance between an anchor and a positive sample, both of which have the same identity, is minimized while the distance between the anchor and a negative sample is maximized. Wen et al. \cite{wen_center_loss_eccv_2016} propose a center loss which only reduces the intra-class variability of features learned with convolutional neural networks. 

Most deep domain adaptation networks contribute by designing novel distance metrics for measuring domain discrepancy and the discriminative representation learning is previously limited to fully supervised settings, for example, face recognition. The effect of discriminative features is unknown in the deep domain adaptation setting, which is the main focus of this paper. The most related work is Scatter Component Analysis (SCA) \cite{ghifary_sca_pami_2016}. Though motivated from a similar intuition, the Fisher loss differs from SCA in the following two aspects: (1) SCA relies on pre-computed features while Fisher loss can be optimized in an end-to-end approach which allows representation learning from raw inputs. (2) SCA only adopts MMD as the transfer criterion. In addition to MMD, the Fisher loss can be used together with CORAL and domain adversarial loss, which is more general. A similar criterion is proposed in \cite{hu_mda_uai_2019} which addresses the domain generalization problem where no target domain data is available during model training, which is different from the domain adaptation setting in this work. 

\section{Fisher Deep Domain Adaptation}
In this section, we first describe the problem setup and present an overview of the deep domain adaptation network with the Fisher loss. Then we introduce each component of the network in detail. 

\subsection{Problem Setup}
We focus on the domain adaptation setting where the source domain is labeled and the target domain is unlabeled. The source domain, denoted by $\mathcal{D}_{s}$, is composed of $n_{s}$ labeled samples $\{\mathbf{x}_{s_j}, y_{s_j}\}_{j=1}^{n_{s}}$ where $\mathbf{x}_{s_j} \in \mathcal{X}_{s}$ and $y_{s_j} \in \mathcal{Y}_{s}$, and $\mathcal{X}_{s}$, $\mathcal{Y}_{s}$ denote the raw input space and the label space of the source domain, respectively. Similarly, the target domain, denoted by $\mathcal{D}_{t}$, contains $n_{t}$ samples $\{\mathbf{x}_{t_j}\}^{n_{t}}_{j=1}$. We limit our discussion to homogeneous classification tasks. Therefore, there is $\mathcal{Y}_{s} = \mathcal{Y}_{t}$. With the subscript omitted, the label space is defined as $\mathcal{Y} = \{1, \ldots, K\}$ where $K$ is the number of classes. Each sample is associated with a domain label, denoted by $d$, which is defined as $d_{s_j} = 0$ for $\mathbf{x}_{s_j} \in \mathcal{D}_{s}$ and $d_{t_j} = 1$ for $\mathbf{x}_{t_j} \in \mathcal{D}_{t}$.

\begin{figure}
    \centering
    \includegraphics[scale=0.75]{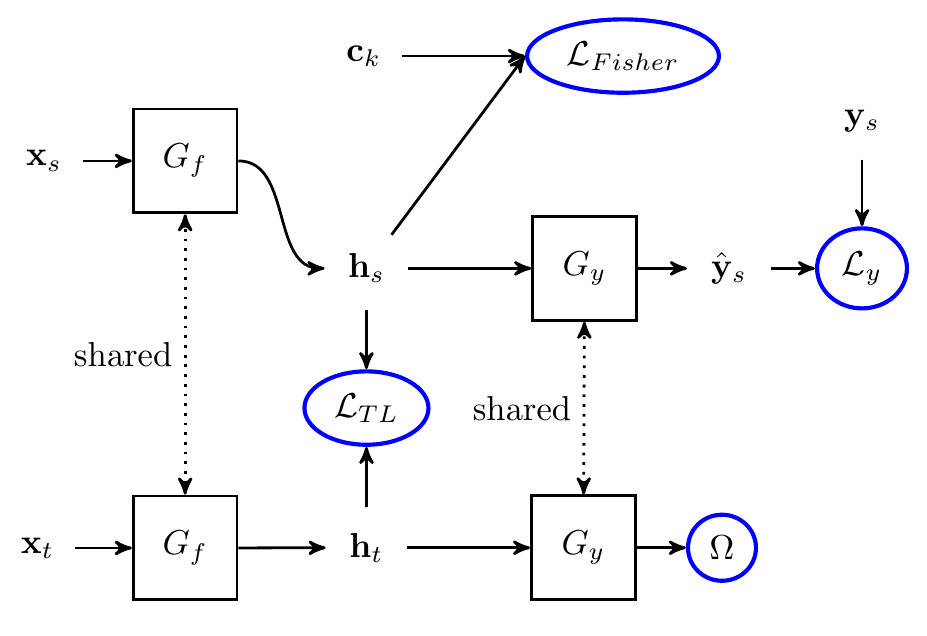}
    \caption{An overview of Fisher Deep Domain Adaptation Network. }
    \label{fig:fisher_dan}

\end{figure}

\subsection{The Objective}
Given the labeled source domain $\mathcal{D}_{s}$ and the unlabeled target domain $\mathcal{D}_{t}$, the objective of a deep domain adaptation network is to learn a functional mapping $g:\mathcal{X} \rightarrow \mathcal{Y}$ for both domains. The functional mapping $g$ can be decomposed into two networks, a feature extractor and a label predictor, denoted by $G_{f}$ and $G_{y}$, respectively. The feature extractor $G_{f}$ maps the raw inputs to a latent feature space. The learned feature representations are denoted by $\mathbf{h}_s$ and $\mathbf{h}_t$, respectively. The label predictor $G_{y}$ then maps the latent features $\mathbf{h}$ to the label space $\mathcal{Y}$ and hence it characterizes the conditional probability distribution $p(y|\mathbf{h})$. Since the target domain is unlabeled, the label predictor $G_y$ can only be trained with the labeled source domain. To reuse the label predictor for the target domain, the distributions $p(\mathbf{h}_s)$ and $p(\mathbf{h}_t)$ need to be aligned. A schematic view is shown in Fig. \ref{fig:fisher_dan} and the overall objective can be defined as
\begin{equation}
 \min_{G_{f}, G_{y}} \quad \mathcal{L}_{y} + \lambda_{0} \cdot \mathcal{L}_{Fisher} + \lambda_{1} \cdot \Omega + \lambda_{2} \cdot \mathcal{L}_{TL} , 
\label{eqn:discriminant-da-overall-loss}
\end{equation}
where $\mathcal{L}_{y}$ is the task loss, $\mathcal{L}_{Fisher}$ denotes the Fisher loss, $\Omega$ denotes the entropy-based regularization term, $\mathcal{L}_{TL}$ is the transfer loss, and $\lambda$'s are regularization parameters to balance the loss terms. For classification tasks, the task loss can be the cross-entropy loss or hinge loss. It provides supervision from the labeled data and produces separable features in the source domain. In addition to the separability, the Fisher loss and the entropy-based regularization further encourages feature discriminability in the labeled source domain and the unlabeled target domain, respectively. The transfer loss $\mathcal{L}_{TL}$ is an empirical estimate of the domain discrepancy. The two distributions $p(\mathbf{h}_s)$ and $p(\mathbf{h}_t)$ are aligned by minimizing the transfer loss. 

\subsection{Fisher Loss}
\label{sec:fisher_loss}
The Fisher loss can be defined as
{\small
\begin{equation}
    \mathcal{L}_{Fisher}=\alpha(\mathrm{tr}(\mathbf{S}_w),\mathrm{tr}(\mathbf{S}_b)),\label{dl_loss}
\end{equation}
}\noindent
where $\mathrm{tr}(\cdot)$ denotes the trace of a square matrix,  $\mathbf{c}_{k}$ is the center of the $k$-th class, $\mathbf{c} = \frac{1}{K} \sum_{k=1}^{K} \mathbf{c}_{k} $ is global center of all classes, $\mathbf{h}_{k,j}$ is the latent feature of the $j$-th sample in class $k$, $\mathbf{S}_{b} = \sum_{k=1}^{K} (\mathbf{c}_{k} - \mathbf{c})(\mathbf{c}_{k} -  \mathbf{c})^{T}$ denotes the between-class scatter matrix, and $\mathbf{S}_{w} = \sum_{k=1}^{K} \sum_{j=1}^{n_{k}} (\mathbf{h}_{k,j} -  \mathbf{c}_{k})(\mathbf{h}_{k,j} -  \mathbf{c}_{k})^{T}$ is the within-class scatter matrix. Note that $\mathbf{c}_{k}$ is a variable to be optimized, which is different from that in LDA. Function $\alpha(\cdot,\cdot)$ in Eq. (\ref{dl_loss}) is required to be monotonically increasing with respect to the first input argument and monotonically decreasing with respect to the second one. Due to this property of function $\alpha(\cdot,\cdot)$, minimizing it in Eq. (\ref{eqn:discriminant-da-overall-loss}) can enforce the within-class compactness via $\mathrm{tr}(\mathbf{S}_w)$ and the between-class separation via $\mathrm{tr}(\mathbf{S}_b)$. The trace function in Eq. (\ref{dl_loss}) is one way to measure the scatter and there are other ways to achieve this, for example, the determinant. Due to the computationally efficient property of the trace function, we adopt it here and other ways are left for future studies.

In the following sections, we will introduce two instantiations of the Fisher loss.

\noindent{\textbf{Trace Ratio. }}
Motivated by Fisher's discriminant analysis, the first instantiation is defined as
{\small
\begin{equation}
    \mathcal{L}_{Fisher-TR} =\mathrm{tr}(\mathbf{S}_{w})/\mathrm{tr}(\mathbf{S}_{b}).
\label{eqn:trace-ratio-minimization}
\end{equation}
}\noindent
Consequently, the corresponding $\alpha(\cdot,\cdot)$ is defined as the ratio function: $\alpha(x_1,x_2)=x_1/x_2$, which obviously satisfies the requirement on $\alpha(\cdot,\cdot)$.

\noindent{\textbf{Trace Difference. }}
Motivated by MMC, the second instantiation is defined as
{\small
\begin{equation}
    \mathcal{L}_{Fisher-TD} = \mathrm{tr}(\mathbf{S}_{w}) - \lambda_{b} \cdot \mathrm{tr}(\mathbf{S}_{b}), 
\label{eqn:mmc}
\end{equation}
}\noindent
where $\lambda_{b}$ is a hyperparameter to control the importance of the between-class penalty and $\alpha(\cdot,\cdot)$ is instantiated as $\alpha(x_1,x_2)=x_1-\lambda_b\cdot x_2$ to satisfy its requirement. The criterion of MMC is a special case of $\mathcal{L}_{Fisher-TD}$ by setting $\lambda_{b}=1$ and hence $\mathcal{L}_{Fisher-TD}$ is a generalization of MMC.

While the Fisher loss is defined as the trace ratio in Eq. (\ref{eqn:trace-ratio-minimization}), Eq. (\ref{eqn:mmc}) defines it as the difference of the trace. The trace difference definition explicitly enforces between-class separability and it is computationally more efficient, but it involves one more hyper-parameter $\lambda_{b}$. 

The features of the source domain samples become discriminative under the guidance of the Fisher loss, but the discriminability is not guaranteed in the target domain because the center assignment unknown and the Fisher loss cannot be applied. To explicitly enforce the discriminability in the target domain, we adopt the entropy minimization regularization \cite{grandvalet_entropy_minimization_nips_2005}, which is formulated as
{\small
\begin{equation}
\begin{aligned}
    \Omega  = - \sum_{j} \sum_{k=1}^{K} p(y_j=k|\mathbf{h}_{j}) \log p(y_j=k|\mathbf{h}_{j}), 
\end{aligned}
\end{equation}
}\noindent
where $p(y|\mathbf{h})$ is estimated with the label predictor. The entropy minimization regularization will push each target domain sample towards one of the class centers. 

\subsection{Transfer Criteria}
\label{sec:transfer-criteria}

Many transfer criteria have been proposed and three widely used criteria, i.e., MMD, CORAL and the domain adversarial loss, are described in the following. 

\noindent{\textbf{Maximum Mean Discrepancy (MMD). }}
The MMD \cite{gretton_mmd_nips_2012} is a non-parametric distance between two probability distributions. It measures the distance of the mean embeddings of the samples from different domains in a Reproducing Kernel Hilbert Space (RKHS). It can be empirically estimated by
{ \small
\begin{equation*}
\begin{aligned}
    \mathcal{L}_{MMD} =& \frac{1}{n_s^{2}} \sum_{i,j}^{n_s} \kappa(\mathbf{h}_{s_i} , \mathbf{h}_{s_j}) + \frac{1}{n_t^{2}} \sum_{i,j}^{n_t} \kappa(\mathbf{h}_{t_i} , \mathbf{h}_{t_j}) \\
    & - \frac{2}{n_s \cdot n_t} \sum_{i,j}^{n_s, n_t} \kappa(\mathbf{h}_{s_i} , \mathbf{h}_{t_j}) , 
\end{aligned}
\end{equation*}
}\noindent
where $\kappa$ is a linear combination of multiple RBF kernels to cover a large range of mean embeddings. 

\noindent{\textbf{Deep Correlation Alignment (CORAL). }}
The CORAL loss \cite{sun_coral_eccv_2016} is defined as the distance of the second-order statistics (covariance) of the latent features from the two domains:
{\small
\begin{equation*}
    \mathcal{L}_{CORAL} = \frac{1}{4p^{2}} \lVert C_{s} - C_{t} \rVert_{F}^{2},
\end{equation*}
}\noindent
where $p$ is the dimension of the latent features and $C_{s}$ and $C_{t}$ are feature covariance matrices of the source domain and the target domain, respectively. 

\noindent{\textbf{Domain Adversarial Loss. }}
The domain discrepancy can be parameterized as a neural network and it is referred to as the domain adversarial loss \cite{ganin_dann_jmlr_2016}. An additional network called discriminator, denoted by $D$, is introduced. It is optimized to predict the domain label of an input latent feature. On the other hand, the feature extractor is optimized with an adversarial objective by confusing the discriminator. When the discriminator fails to distinguish latent features from the two domains, the shared feature space is found. The domain adversarial loss is defined as: 
{\small
\begin{equation*}
\begin{aligned}
\mathcal{L}_{adv}=& \frac{1}{n_s} \sum_{\mathbf{x}_s \in \mathcal{D}_{s}} [ \ell_{d}(D(G_f(\mathbf{x}_s)), d_{s})] \\
& +\frac{1}{n_t} \sum_{\mathbf{x}_t \in \mathcal{D}_{t}} [\ell_{d}(D(G_f(\mathbf{x}_t)), d_{t})].   
\end{aligned}
\end{equation*}
}\noindent
Hence the transfer criterion here is $\mathcal{L}_{TL} = \max_D\{-\mathcal{L}_{adv}\}$. 

\subsection{Optimization}
\label{sec:fisher_loss_optimization}
Since the other parameters in the network can be optimized with typical stochastic gradient descent, we focus on optimizing the Fisher loss in this section. Given the definitions of the scatter matrices, the estimation of the scatter matrices involves estimating the class centers $\{\mathbf{c}_{k}\}_{k=1}^{K}$. To attain an exact estimation of the centers, it is necessary to go through the entire training set which can not be done on a mini-batch basis. Instead of an exact estimation, the class centers are treated as variables so that they are optimized jointly with the other parameters in the network \cite{wen_center_loss_eccv_2016}. 

Based on the definition of matrix norm and trace, the within-class scatter matrix can be reformulated as
{\small
\begin{equation}
    \mathrm{tr}(\mathbf{S}_{w}) = \sum_{j=1}^{m} \lVert \mathbf{h}_{j} - \mathbf{c}_{y_j} \rVert_{F}^{2}, 
\label{eqn:intra-class-loss}
\end{equation}
}\noindent
where $m$ is the mini-batch size and $\lVert \cdot \rVert_{F}$ is the Frobenius norm. Similarly, the between-class scatter matrix can be estimated by the distance between the class centers to the global center as
{\small
\begin{equation}
    \mathrm{tr}(\mathbf{S}_{b}) = \sum_{k=1}^{K} \lVert \mathbf{c}_{k} - \mathbf{c} \rVert_{F}^{2}.
\label{eqn:inter-class-loss-center}
\end{equation}
}

With Eqs. (\ref{eqn:intra-class-loss}) and (\ref{eqn:inter-class-loss-center}), the gradient of the discriminative loss defined in Eq. (\ref{eqn:trace-ratio-minimization}) with respect to the feature extractor and the class centers are
{\small
\begin{equation*}
    \begin{aligned}
    \frac{\partial \mathcal{L}_{Fisher-TR}}{\partial \mathbf{h}_{j}} =& \frac{2 \cdot (\mathbf{h}_{j} - \mathbf{c}_{y_j})}{\mathrm{tr}(\mathbf{S}_{b})}, \\
    \frac{\partial \mathcal{L}_{Fisher-TR}}{\partial \mathbf{c}_{k}} =& \frac{\sum_{j=1}^{m} 2 \cdot \mathbb{I}(y_j = k) \cdot (\mathbf{c}_{k} - \mathbf{h}_{j})}{\mathrm{tr}(\mathbf{S}_{b})} \\
    & - \frac{2 (1-\frac{1}{K}) \cdot \mathrm{tr}(\mathbf{S}_{w})}{\mathrm{tr}(\mathbf{S}_{b})^{2}} (\mathbf{c}_{k} - \frac{1}{K} \sum_{j=1}^{K} \mathbf{c}_{j}),
    \end{aligned}
\label{eqn:grad-trace-ratio-minimization}
\end{equation*}
}\noindent
where $\mathbb{I}(\cdot)$ denotes the indicator function. 

Similarly, when the Fisher loss adopts the form of the trace difference in Eq. (\ref{eqn:mmc}), the update rules are computed as 
{\small
\begin{equation*}
    \begin{aligned}
    \frac{\partial \mathcal{L}_{Fisher-TD}}{\partial \mathbf{h}_{j}} &= 2 \cdot (\mathbf{h}_{j} - \mathbf{c}_{y_j}) , \\
    \frac{\partial \mathcal{L}_{Fisher-TD}}{\partial \mathbf{c}_{k}} &= \sum_{j=1}^{m} 2 \cdot \mathbb{I}(y_j = k) \cdot (\mathbf{c}_{k} - \mathbf{h}_{j}) \\
    & - 2 (1-\frac{1}{K}) \cdot (\mathbf{c}_{k} - \frac{1}{K} \sum_{j=1}^{K} \mathbf{c}_{j}).  
    \end{aligned}
\label{eqn:grad-mmc}
\end{equation*}
}\noindent

\subsection{Theoretical Insights}
\label{sec:theoretical-insights}
A theoretical explanation for the proposed Fisher loss is provided in the following. 
\begin{theorem}(Generalization bound of domain adaptation  \cite{ben_da_theory_nips_2007,ganin_dann_jmlr_2016})
\label{th:da_theory}
Given a fixed representation mapping, let $\mathcal{H}$ denote a hypothesis space, $\epsilon_{s}(\eta)$ and $\epsilon_{t}(\eta)$ denote the empirical risk of a hypothesis $\eta$ in the source domain and target domain, respectively. For arbitrary $\eta \in \mathcal{H}$, with probability at least $1-\delta$, we have
{\small
\begin{equation*}
    \epsilon_{t}(\eta) \leq \epsilon_{s}(\eta) + d_{\mathcal{H}}(p(\mathbf{h}_{s}), p(\mathbf{h}_{t})) + \beta + C, 
\end{equation*}
}\noindent
where $d_{\mathcal{H}}(p(\mathbf{h}_{s}), p(\mathbf{h}_{t}))$ is the $\mathcal{H}$-divergence between the probability distributions of the source domain and the target domain, $\beta$ denotes the empirical risk of a hypothesis on both domains with $\beta \geq \inf_{\eta^{*} \in \mathcal{H}} [\epsilon_{s}(\eta^{*}) + \epsilon_{t}(\eta^{*})]$, and $C$ denotes a constant term. 
\end{theorem}

The transfer loss is an empirical estimate of the $\mathcal{H}$-divergence. It is expected that in the learned latent feature space there is an ideal hypothesis with low risks on both domains, which corresponds to a low $\beta$ value in Theorem \ref{th:da_theory}. However, such a hypothesis might not exist without the discriminability constraint, as illustrated by the toy example in Fig. \ref{fig:toy_example}. On the contrary, the Fisher loss encourages the discriminability in the source domain. Meanwhile, the transfer loss and the entropy-based regularization strengthen the discriminability in the target domain. When the features on both domains are discriminative, a hypothesis that performs well on both domains is more likely to be found. With a small $\beta$, a tight generalization bound of $\epsilon_{t}(\eta)$ can be achieved. 

\section{Experiments}


In this section, we conduct empirical studies to test the proposed Fisher loss. Code will be available at \url{https://github.com/HKUST-KnowComp/FisherDA}. The proposed model is evaluated on two domain adaptation benchmarks, \textit{Office-31} \cite{saenko_office_eccv_2010} and \textit{Office-Home} \cite{venkateswara_office_home_cvpr_2017}, whose details are introduced as follows.
\begin{enumerate}
\item \textit{Office-31}. There are $3$ domains in the \textit{Office-31} dataset: Amazon (A), Webcam (W) and DSLR (D), and $6$ transfer tasks can be constructed. There are $31$ classes in each domain. 
\item \textit{Office-Home}. There are $4$ domains in the \textit{Office-Home} dataset: Art (A), Clipart (C), Product (P) and Real World (R), and $12$ transfer tasks can be constructed. There are $65$ classes in each domain. 
\end{enumerate}

The classification accuracy is adopted as the evaluation metric. Four baseline models and their counterparts which use the Fisher loss are evaluated. The four baseline models are: \textit{Source} where the model is merely trained in the source domain and then directly applied to the target domain without any adaptation; \textit{MMD}, \textit{CORAL} and \textit{ADA} where MMD, CORAL and domain adversarial loss are used as the transfer loss, respectively. In addition, the results of four state-of-the-art deep domain adaptations models, Deep Adaptation Network (DAN) \cite{long_dan_icml_2015}, Residual Transfer Network (RTN) \cite{long_rtn_nips_2016}, Joint Adaptation Network (JAN) \cite{long_jan_icml_2017} and Multi-Adversarial Domain Adaptation (MADA) \cite{pei_mada_aaai_2018}, are reported on the \textit{Office-31} dataset. 

All the models are implemented with the PyTorch framework. \textit{ResNet-50} \cite{he_resnet_cvpr_2016} is used as the backbone network. The network is initialized with the parameters which are pre-trained on the ImageNet dataset \cite{ILSVRC_ijcv_2015}. An additional fully-connected layer with $256$ units are stacked on top of the backbone network as part of the feature extractor and all the parameters are fine-tuned during the training process. The neural networks are optimized with the mini-batch stochastic gradient descent with the momentum of $0.9$. Early stopping is used, that is, if the network performance does not improve within $2500$ mini-batches, the training process is terminated. The batch size $m$ equals to $36$ and weight decay is $0.0005$ in the experiment. Hyperparameters are selected with the cross validation method proposed in \cite{sugiyama_iwcv_jmlr_2007}. For the \textit{MMD} model, a linear combination of $19$ RBF kernels is used. The standard deviation parameters of the RBF kernels are $\{10^{-6}, 10^{-5},10^{-4},10^{-3},10^{-2},10^{-1},1,5,10,15,20,25,\allowbreak30,35,100,10^3,10^{4},10^{5}, 10^{6}\}$ \cite{bousmalis_dsn_nips_2016}. 
The learning rate is selected from $\{0.001, 0.0003\}$. The learning rate of the parameters of the bottleneck layer, the classification layer, the centers and the domain discriminators are $10$ times that of the other parts of the network as they are learned from scratch. The learning rate is decayed with
\begin{equation*}
\small
\eta = \frac{\eta_0}{(1+ \omega \cdot p)^{\rho}},
\end{equation*}
where $p$ indicates the training progress, $\eta_0$ is the initial learning rate, and $\omega$ and $\rho$ are hyperparameters. 

For the Fisher loss in the trace difference form, the weight $\lambda_0$ is selected from $\{10^{-3}, 10^{-4}\}$ and the weight $\lambda_b$ is selected from $\{0, 0.5, 1, 5, 10\}$. For the Fisher loss in the trace ratio form, the weight $\lambda_0$ is selected from $\{0.1, 1\}$. The weight of the entropy minimization regularization term $\lambda_1$ is selected from $\{0.0, 0.1\}$. The weight of the transfer loss $\lambda_2$ is selected from $\{0.1, 1.0, 10.0\}$. For adversarial learning models, the trade-off parameter $\lambda_{2}$ of the adversarial loss $\mathcal{L}_{adv}$ is scheduled as defined in \cite{ganin_dann_jmlr_2016}: 
\begin{equation*}
\small
\lambda_{2} = \frac{2 \cdot u}{1+ \exp(-\gamma \cdot p) } - u,
\end{equation*}
where $u$ is the upper bound of the trade-off parameter and $\gamma$ is a hyperparameter. The trade-off parameter $\lambda_{2}$ increases from $0$ to $u$, which avoids noisy signals at the early stage of training. 

\subsection{Experimental Results}

\begin{figure*}[!htbp]
    \centering
    \subfloat[Office-31] {
    \begin{minipage}{0.48\linewidth}
    \centering
    \includegraphics[scale=0.45]{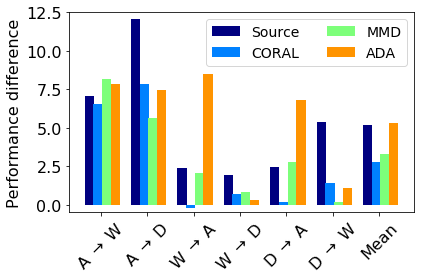}
    \end{minipage}
    }
    \hfill
    \subfloat[Office-Home] {
    \begin{minipage}{0.48\linewidth}
    \centering
    \includegraphics[scale=0.45]{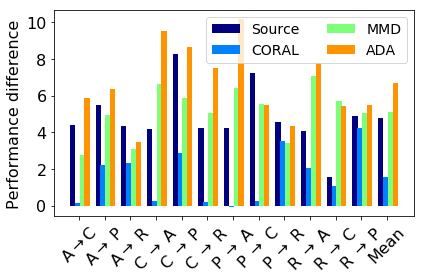}
    \end{minipage}
    }
    \caption{Performance difference brought by the proposed Fisher loss on two benchmarks.}
    \label{fig:abs_bar}
\end{figure*}

Results on the two datasets are shown in Tables \ref{tab:office} and \ref{tab:office-home}, respectively. The accuracy is averaged over two random trials. The optimal performance of each transfer task is highlighted in bold face. The performance difference of each model with the Fisher loss over that without the Fisher loss on different domains is shown in Fig. \ref{fig:abs_bar}. Since there are two instantiations of the Fisher loss, the higher accuracy out of the two instantiations is used for calculating the performance difference. 

\begin{table*}[!htbp]
\centering
\caption{Classification accuracies on the \textit{Office-31} dataset.}
\label{tab:office}
\resizebox{\textwidth}{!}{
\begin{tabular}{ccccccccc}
\toprule
 &  & A $\rightarrow$ W & A $\rightarrow$ D & W $\rightarrow$ A & W $\rightarrow$ D & D $\rightarrow$ A & D $\rightarrow$ W & Mean \\
 \midrule
\multirow{4}{*}{Baselines} & Source & 65.53 & 72.39 & 58.25 & 98.09  & 60.24 & 93.65 & 74.69 \\
 & CORAL  & 71.38 & 77.11 & 62.71 & 99.20  & 63.70 & 97.30 & 78.57 \\
 & MMD & 85.28 & 83.13 & 63.74 & 99.20  & 64.48 & 97.61 & 82.24 \\
 & ADA   & 85.28 & 82.63 & 61.43 & 99.50  & 64.80 & 97.48 & 81.85 \\
 \cline{2-9}
 & DAN \cite{long_dan_icml_2015} & 80.5 & 78.6 & 62.8 & 99.6 & 63.6 & 97.1 & 80.4 \\
 & RTN \cite{long_rtn_nips_2016} & 84.5 & 	77.5 & 64.8 & 99.4 & 66.2 & 96.8 & 81.5 \\
 & JAN \cite{long_jan_icml_2017} & 86.0 & 85.1 & \textbf{70.7} & 99.8 & 69.2 & 97.4 & 84.7 \\
 & MADA \cite{pei_mada_aaai_2018} & 90.0 & 87.8 & 66.4 & 99.6 & 70.3 & 97.4 & 85.3 \\
 \midrule
\multirow{4}{*}{w/ $\mathcal{L}_{Fisher-TD}$} & Source & 72.58 & 84.44 & 59.18 & 99.60  & 62.67 & 97.99 & 79.41 \\
 & CORAL  & 77.92 & 83.84 & 62.50 & 99.90  & 63.06 & 98.74 & 80.99 \\
 & MMD & 93.21 & 88.76 & 64.15 & 99.90  & 67.23 & 97.80 & 85.17 \\
 & ADA  & 93.14 & 89.16 & 67.64 & 99.80  & \textbf{71.60} & 98.24 & 86.60 \\
 \midrule
\multirow{4}{*}{w/ $\mathcal{L}_{Fisher-TR}$} & Source & 72.01 & 83.53 & 60.63 & \textbf{100.00} & 61.52 & \textbf{98.99} & 79.45 \\

 & CORAL  & 77.74 & 84.94 & 61.79 & 99.90  & 63.88 & 98.49 & 81.12 \\
 & MMD & \textbf{93.46} & 87.85 & 65.83 & \textbf{100.00} & 66.86 & 97.80 & 85.30 \\
 & ADA  & 92.77 & \textbf{90.06} & 69.90 & 99.80  & 71.42 & 98.55 & \textbf{87.08} \\
\bottomrule
\end{tabular}
}
\end{table*}

\begin{table*}[!htbp]
\vskip -0.1in
\centering
\caption{Classification accuracies on the \textit{Office-Home} dataset.}
\label{tab:office-home}
\resizebox{\textwidth}{!}{
\begin{tabular}{ccccccccccccccc}
\toprule
& & A$\rightarrow$C & A$\rightarrow$ P & A$\rightarrow$ R & C $\rightarrow$ A & C $\rightarrow$ P & C $\rightarrow$ R & P $\rightarrow$ A & P $\rightarrow$ C & P $\rightarrow$ R & R $\rightarrow$ A & R $\rightarrow$ C & R $\rightarrow$ P & Mean \\
\midrule
\multirow{4}{*}{Baselines} & Source & 36.39 & 59.34 & 70.19 & 40.94 & 54.04 & 57.03 & 44.89 & 30.00 & 68.63 & 60.55 & 39.62 & 72.73 & 52.86 \\
& CORAL  & 41.16 & 62.45 & 71.61 & 47.01 & 58.35 & 61.18 & 48.56 & 36.68 & 70.40 & 62.92 & 43.10 & 72.76 & 56.35 \\
& MMD    & 41.95 & 62.15 & 71.64 & 46.23 & 59.54 & 61.69 & 46.66 & 38.55 & 70.05 & 61.99 & 47.22 & 74.49 & 56.85 \\
& ADA    & 43.97 & 60.05 & 70.34 & 46.70 & 57.31 & 59.73 & 45.94 & 43.33 & 69.34 & 62.11 & 52.92 & 75.34 & 57.26 \\
\midrule
\multirow{4}{*}{w/ $\mathcal{L}_{Fisher-TD}$} & Source & 40.77 & 62.15 & 71.12 & 41.74 & 58.63 & 60.86 & 49.13 & 37.22 & 70.08 & 64.44 & 37.06 & 72.29 & 55.46 \\
& CORAL  & 39.79 & 61.87 & 70.90 & 47.28 & 58.04 & 60.86 & 48.48 & 35.95 & 70.20 & 64.38 & 40.54 & 73.10 & 55.95 \\
& MMD    & 43.91 & 66.92 & \textbf{74.73} & 52.84 & 64.95 & 66.35 & 53.05 & 43.26 & 73.48 & 68.21 & 52.62 & 79.53 & 61.65 \\
& ADA    & \textbf{49.86} & 66.40 & 73.80 & 55.79 & \textbf{65.97} & 65.94 & 55.54 & \textbf{48.80} & 73.70 & \textbf{69.86} & 57.87 & 80.77 & \textbf{63.69} \\
\midrule
\multirow{4}{*}{w/ $\mathcal{L}_{Fisher-TR}$} & Source & 39.51 & 64.83 & 74.52 & 45.12 & 62.31 & 61.28 & 42.97 & 30.63 & 73.16 & 64.61 & 41.18 & 77.62 & 56.48 \\
& CORAL  & 41.27 & 64.63 & 73.95 & 47.14 & 61.23 & 61.37 & 45.49 & 36.95 & \textbf{73.90} & 64.96 & 44.17 & 76.98 & 57.67 \\
& MMD   & 44.72 & \textbf{67.08} & 74.18 & 52.68 & 65.43 & 66.73 & 52.64 & 44.09 & 73.05 & 69.06 & 52.92 & 79.06 & 61.80 \\
& ADA   & 49.62 & 65.50 & 73.80 & \textbf{56.24} & 63.37 & \textbf{67.25} & \textbf{56.10} & 47.87 & 73.19 & 67.90 & \textbf{58.35} & \textbf{80.82} & 63.33 \\
\bottomrule
\end{tabular}
}
\vskip -0.15in
\end{table*}

Without the Fisher loss, the naive transfer model \textit{Source} performs the worst among all models on all domains, which indicates that there is a non-negligible discrepancy between the source domain and the target domain. Therefore, using a transfer criterion becomes necessary. Overall, the adaptation performance is improved with all the transfer criteria. The \textit{CORAL} model slightly surpasses the \textit{Source} baseline. \textit{MMD} and \textit{ADA} achieve comparable results and they both exceed the other models with an obvious margin. This demonstrates the efficacy of the transfer criteria.

Models with the Fisher loss outperform their counterparts with a noticeable margin. The absolute improvement in terms of mean accuracy of the $6$ transfer tasks on the \textit{Office-31} are $5.19\%$, $2.75\%$, $3.27\%$ and $5.32\%$, respectively. The performances of the two instantiations of the Fisher loss are comparable. The \textit{ADA} model with the Fisher loss surpasses state-of-the-art deep domain adaptation baselines such as JAN and MADA and it achieves the optimal performance on $5$ out of $6$ tasks. Hence, with the Fisher loss, a traditional transfer criterion can achieve comparable results as advanced transfer criteria, which demonstrates the effectiveness of the Fisher loss. 
Even for the \textit{Source} model without the use of a transfer criterion, the mean accuracy on the \textit{Office-31} dataset is increased from $74.69\%$ to $79.89\%$, which is comparable to the \textit{CORAL} model without the Fisher loss. It indicates that the transfer criterion and the Fisher loss are complementary to each other. 

A similar tendency is observed on the \textit{Office-Home} dataset. The performance gain is even more obvious since the number of classes of \textit{Office-Home} is twice as much as that of \textit{Office-31}. The absolute improvement can be up to $10.16\%$, which is obtained by the \textit{ADA} network on the P $\rightarrow$ A task. Together with the Fisher loss, the network with the domain adversarial loss has an apparent advantage over other transfer criteria. This might be because the Fisher loss can fully unleash the potential of the ``network'' distance, which is used in the domain adversarial loss, for deep domain adaptation.

\begin{figure*}[!thb]
    \centering
    \subfloat[The \textit{ADA} model] {
    \begin{minipage}[c]{0.48\linewidth}
    \centering
    \includegraphics[scale=0.3, trim=0 1em 0 0, clip]{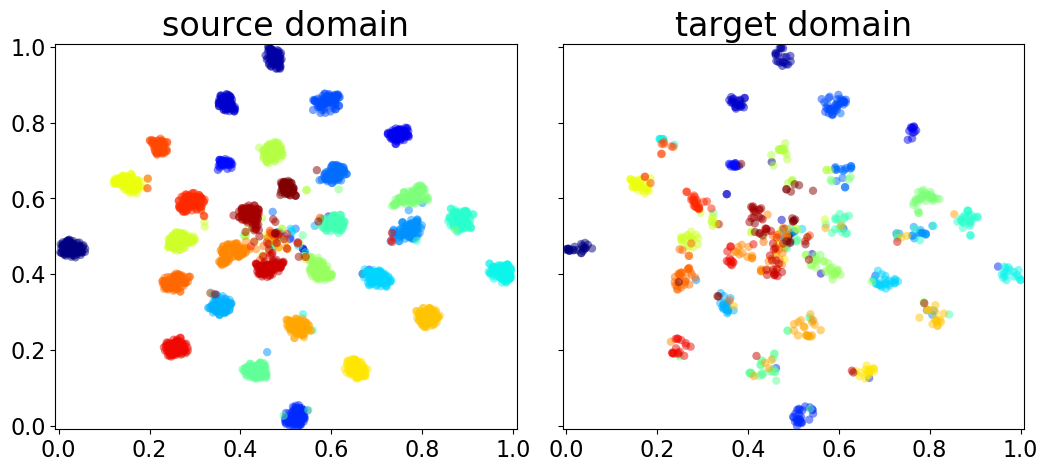}
    \label{fig:native_dann_class}
    \end{minipage}
    }
    \hfill
    \subfloat[w/ $\mathcal{L}_{Fisher-TR}$ and entropy-based regularization] {
    \begin{minipage}[c]{0.48\linewidth}
    \centering
    \includegraphics[scale=0.3, trim=0 1em 0 0, clip]{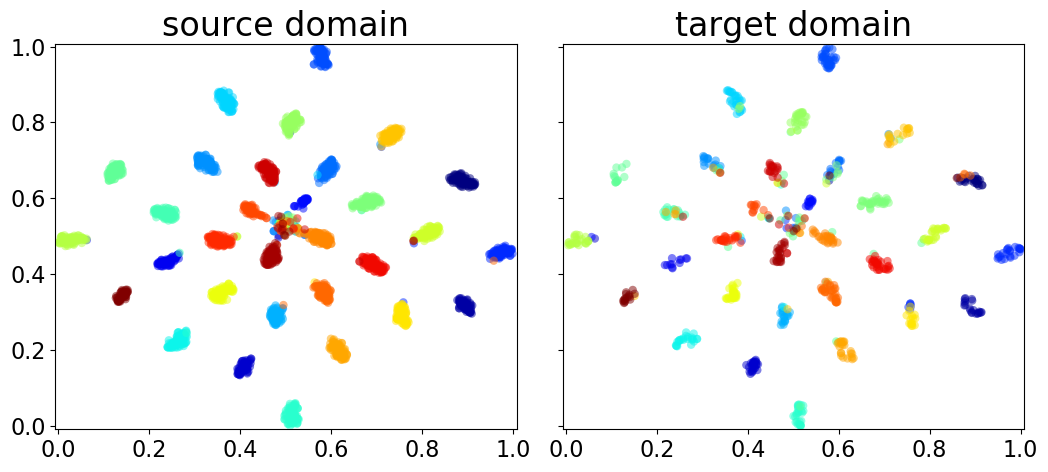}
    \label{fig:dann_tr_wo_ent_tsne}
    \end{minipage}
    }

    \caption{The probability distributions of the source domain (Amazon) and the target domain (Webcam) in the latent feature space produced by \textit{ADA} model variants (The samples with the same class label are plotted with the same color. ). The latent representations become more discriminative with the proposed Fisher loss and entropy-based regularization than they are with the base \textit{ADA} model and hence better adaptation performance is achieved. }
    \label{fig:tsne}
    \vskip -0.1in
\end{figure*}

\subsection{Ablation Analysis}

To provide further insights into the proposed Fisher loss, multiple model variants are explored and the learned latent representations are visualized. The experiments are conducted on the A $\rightarrow$ W setting of the \textit{Office-31} dataset and the \textit{ADA} model is used as the base model.

\begin{table}[htbp]
\vskip -0.1in
    \centering
    \caption{Classification accuracy of the \textit{ADA} model variants ($\checkmark$/ \xmark\, denote that the loss term is used/not used in the model)}
    \label{tab:dann-ablation}
    \begin{tabular}{ccc}
    \toprule
        Fisher loss & Entropy minimization & Accuracy \\
        \midrule
        \xmark & \xmark & $85.28$ \\
        \xmark & \checkmark & $90.19$ \\
        \midrule
        Trace Difference & \xmark & $91.70$ \\
        Trace Difference & \checkmark & $93.14$ \\
        \midrule
        Trace Ratio & \xmark & $90.69$ \\
        Trace Ratio & \checkmark & $92.77$ \\
    \bottomrule
    \end{tabular}
\end{table}

The classification accuracy of model variants is shown in Table \ref{tab:dann-ablation} and their t-SNE \cite{maaten_tsne_jmlr_2008} visualizations are shown in Fig. \ref{fig:tsne}. The baseline model \textit{ADA} that corresponds to the first row in Table \ref{tab:dann-ablation} achieves an accuracy of $85.28\%$. With the introduction of the entropy minimization regularization, an apparent improvement of $4.91\%$ is obtained. For the Fisher loss in the form of the trace difference, an absolute improvement of $6.42\%$ is attained though there is no entropy minimization regularization. With the entropy minimization regularization on top of the $\mathcal{L}_{Fisher-TD}$, an additional improvement of $1.44\%$ can be achieved. This is because the entropy minimization regularization further strengthen the feature discriminability in the target domain. For the Fisher loss in the form of the trace ratio, its adaptation performance is comparable to the one in the form of the trace difference. From the visualization in Fig. \ref{fig:tsne}, it is found that the latent representations become more discriminative with the proposed Fisher loss and entropy-based regularization than they are with the base \textit{ADA} model. The ablation study justifies that both the Fisher loss and the entropy minimization are important to the adaptation performance. 

\begin{figure*}
    \centering
    \begin{minipage}{.35\linewidth}
    \centering
    \includegraphics[scale=0.35]{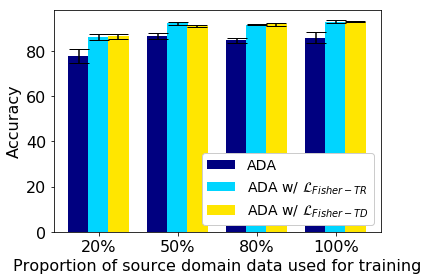}
    \caption{The effect of the number of samples in the source domain. }
    \label{fig:ratio_source_domain}
    \end{minipage}
    \hfill
    \begin{minipage}{.3\linewidth}
    \centering
    \includegraphics[scale=0.32]{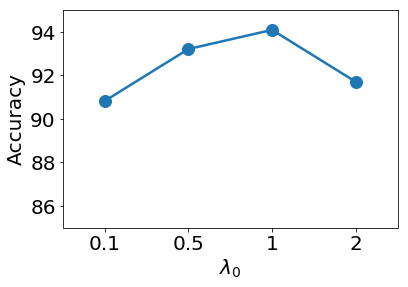}
    \caption{Parameter sensitivity of $\mathcal{L}_{Fisher-TR}$. }
    \label{fig:sensitivity_TR}
    \end{minipage}
    \hfill
    \begin{minipage}{.3\linewidth}
    \centering
    \includegraphics[scale=0.32]{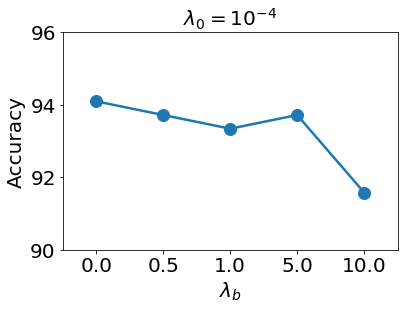}
    \caption{Parameter sensitivity of $\mathcal{L}_{Fisher-TD}$. }
    \label{fig:sensitivity_TD}
    \end{minipage}
\end{figure*}

The effect of the number of samples in the source domain over the adaptation performance on the target domain is studied and the results are shown in Fig. \ref{fig:ratio_source_domain}. We randomly sample $20\%$, $50\%$, $80\%$ and $100\%$ data from the source domain, learn domain adaptation models with the sampled data, and evaluate the adaptation performance. The models with the Fisher loss consistently outperform the base \textit{ADA} model. Though the adaptation performance deteriorates as the number of source domain samples reduces, the accuracy of the models with the Fisher loss is no lower than $86\%$ and hence very stable. On the contrary, the accuracy of the base \textit{ADA} model drops below $80\%$ when only $20\%$ of source domain data are available. With only $20\%$ of the source domain data, the models with the Fisher loss even perform comparably with the base \textit{ADA} model which is trained with the whole source domain. 
The Fisher loss introduces additional hyperparameters. That is, there is one hyperparameter $\lambda_0$ for the Fisher loss in the form of the trace ratio and there is an additional hyperparameter $\lambda_b$ for the Fisher loss in the form of the trace difference. For the Fisher loss in the form of the trace difference, we set $\lambda_0$ to $10^{-4}$ and study the effect of $\lambda_b$. The adaptation performances w.r.t various hyperparameter values are shown in Figs. \ref{fig:sensitivity_TR} and \ref{fig:sensitivity_TD}. The accuracy constantly exceeds $90\%$, which demonstrates that the Fisher loss is not very sensitive to a wide range of parameter values and so makes the selection of hyperparameters easier. 

\section{Conclusion}

The latent features produced by deep domain adaptation models are usually separable while not discriminative, which hampers the adaptation performance. To fill the gap, we propose a Fisher loss to enforce feature discriminability. It reduces the feature variance and hence the resultant hypothesis is more likely to generalize to the target domain. Experimental results demonstrate that the Fisher loss is a general and effective loss for deep domain adaptation as it boosts the performance of three widely adopted transfer criteria. In the future work, we will apply the Fisher loss to improve other transfer criteria. 

\section*{Acknowledgements}
The authors would like to thank Guangneng Hu and Zheng Li for helpful discussions. 

\bibliographystyle{plain}
\bibliography{reference}

\end{document}